\RequirePackage{fix-cm}

\documentclass{svjour3}                     
\smartqed  
\usepackage{graphicx}
\usepackage{multirow}
\usepackage{longtable}
\usepackage{bbding}
\usepackage{array, booktabs, makecell}

\usepackage{latexsym}
\def\Fig#1{Fig. ~\ref{fig:#1}}

\def\Tab#1{Table~\ref{tab:#1}}
\def\Sec#1{Section~\ref{sec:#1}}

\journalname{"Journal of Medical Systems"}
\begin{document}

\title{Medical Image Analysis using Convolutional Neural Networks: A Review}

\author{{Syed Muhammad Anwar$^{1}$} \and Muhammad Majid$^{2}$ \and Adnan Qayyum$^{2}$ \and Muhammad Awais$^{3}$ \and Majdi Alnowami$^{4}$ \and Muhammad Khurram Khan$^{5}$}


\institute{{Syed Muhammad Anwar} \\\\
              {\Envelope \space Muhammad Majid} \\
              {m.majid@uettaxila.edu.pk} \\ \\
              {\space Adnan Qayyum} \\\\
              {\space Muhammad Awais} \\\\
           {\space Majdi Alnowami} \\\\
              {\space Muhammad Khurram Khan} \\\\
           {$^{1}$ Department of Software Engineering, University
of Engineering and Technology Taxila, 47050, Pakistan.}\\
		   {$^{2}$ Department of Computer Engineering, University
of Engineering and Technology Taxila, 47050, Pakistan.}\\
{$^{3}$ Centre for Vision, Speech and Signal Processing (CVSSP), University of Surrey, UK.}\\
{$^{4}$ Department of Nuclear Engineering, King Abdul Aziz University, Jeddah, Saudi Arabia.}\\
{$^{5}$ Center of Excellence in Information Assurance (CoEIA), King Saud University, Riyadh, 11653, Saudi Arabia.}}

\date{Received: date / Accepted: date}

\maketitle

\begin{abstract}
The science of solving clinical problems by analyzing images generated in clinical practice is known as medical image analysis. The aim is to extract information in an affective and efficient manner for improved clinical diagnosis. The recent advances in the field of biomedical engineering has made medical image analysis one of the top research and development area. One of the reason for this advancement is the application of machine learning techniques for the analysis of medical images. Deep learning is successfully used as a tool for machine learning, where a neural network is capable of automatically learning features. This is in contrast to those methods where traditionally hand crafted features are used. The selection and calculation of these features is a challenging task. Among deep learning techniques, deep convolutional networks are actively used for the purpose of medical image analysis. This include application areas such as segmentation, abnormality detection, disease classification, computer aided diagnosis and retrieval. In this study, a comprehensive review of the current state-of-the-art in medical image analysis using deep convolutional networks is presented. The challenges and potential of these techniques are also highlighted. 
             
\end{abstract}

\section{Introduction}
Deep learning (DL) is a widely used tool in research domains such as computer vision, speech analysis, and natural language processing (NLP). This method is suited particularly to those areas, where a large amount of data needs to be analyzed and human like intelligence is required. The use of deep learning as a machine learning and pattern recognition tool is also becoming an important aspect in the field of medical image analysis. This is evident from the recent special issue on this topic \cite{ref1}, where the initial impact of deep learning in the medical imaging domain is investigated. According to an MIT technological review, deep learning is among the top ten breakthroughs of 2013 \cite{ref2}. Medical imaging has been a diagnostic method in clinical practices for a long time. The recent advancements in hardware design, safety procedures, computational resources and data storage capabilities have greatly benefited the field of medical imaging. Currently, major application areas of medical image analysis involve segmentation, classification, and abnormality detection using images generated from a wide spectrum of clinical imaging modalities.  

Medical image analysis aims to aid radiologist and clinicians to make diagnostic and treatment process more efficient. The computer aided detection (CADx) and computer aided diagnosis (CAD) relies on effective medical image analysis making it crucial in terms of performance, since it would directly affect the process of clinical diagnosis and treatment \cite{refMS7,refMS8}. Therefore, the performance of important prameters such as accuracy, F-measure, precision, recall, sensitivity, and specificity is crucial, and it is mostly desirable that these measures give high values in medical image analysis. As the availability of digital images dealing with clinical information is growing, therefore a method that is best suited to big data analysis is required. The state-of-the-art in data centric areas such as computer vision shows that deep learning methods could be the most suitable candidate for this purpose. Deep learning mimics the working of the human brain \cite{ref4}, with a deep architecture composed of multiple layers of transformations. This is similar to the way information is processed in the human brain \cite{ref5}. 

A good knowledge of the underlying features in a data collection is required to extract the most relevant features. This could become tedious and difficult when a huge collection of data needs to be handled efficiently. A major advantage of using deep learning methods is their inherent capability, which allows learning complex features directly from the raw data. This allows us to define a system that does not rely on hand-crafted features, which are mostly required in other machine learning techniques. These properties have attracted attention for exploring the benefits of using deep learning in medical image analysis. The future of medical applications can benefit from the recent advances in deep learning techniques. There are multiple DL open source platforms available such as caffe, tensorflow, theano, keras and torch to name a few \cite{shi2016benchmarking}. The challenges arise due to limited clinical knowledge of DL experts and limited DL knowledge of clinical experts. A recent tutorial attempts to bridge this gap by providing a step by step implementation detail of applying DL to digital pathology images \cite{janowczyk2016deep}. In \cite{lakhani2018hello}, a high-level introduction to medical image segmentation task using deep learning is presented by providing the code. In general, most of the work using DL techniques use an open source model, where the code is made available on platforms such as github. This allows researchers to come up with a running model relatively quickly for applying these techniques to various medical image analysis tasks. The challenge remains to select an appropriate DL architecture depending upon the number of available images and ground truth labels.

\begin{figure}[t]
\begin{center}
\begin{tabular}{c}
\includegraphics[width=110mm]{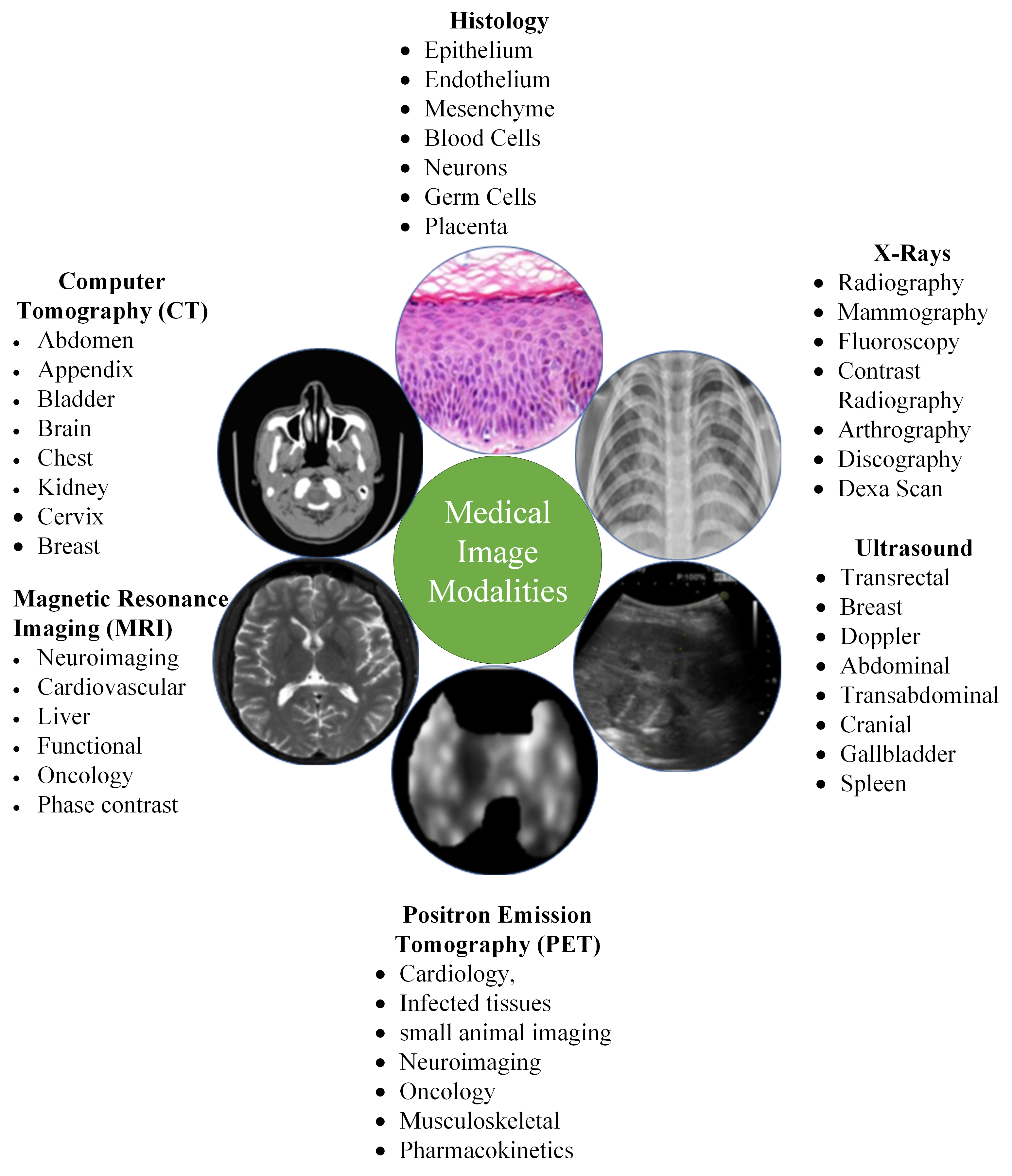}
\end{tabular}
\end{center}
\caption
{ \label{fig:fig1}
{Typology of medical imaging modalities.}}
\end{figure}

In this paper, a detailed review of the current state-of-the-art medical image analysis techniques is presented, which are based on deep convolutional neural networks. A summary of the key performance parameters having clinical significance achieved using deep learning methods is also discussed. The rest of the paper is organized as follows. \Sec{mia}, presents a brief introduction to the field of medical image analysis. \Sec{dl} and Section 4, presents a summary and applications of the deep convolutional neural network methods to medical image analysis. In Section 5, the recent advances in deep learning methods for medical image analysis are analyzed. This is followed by the conclusions presented in \Sec{conc}.

\section{Medical Image Analysis}\label{sec:mia}
Medical imaging includes those processes that provide visual information of the human body. The purpose of medical imaging is to aid radiologists and clinicians to make the diagnostic and treatment process more efficient. Medical imaging is a predominant part of diagnosis and treatment of diseases and represent different imaging modalities. These include X-ray, computed tomography (CT), magnetic resonance imaging (MRI), positron emission tomography (PET), and ultrasound to name a few as well as hybrid modalities \cite{ref7}. These modalities play a vital role in the detection of anatomical and functional information about different body organs for diagnosis as well as for research \cite{ref8}. A typology of common medical imaging modalities used for different body parts which are generated in radiology and laboratory settings is shown in \Fig{fig1}. Medical imaging is an essential aid in modern healthcare systems. Machine learning plays a vital role in CADx with its applications in tumor segmentation, cancer detection, classification, image guided therapy, medical image annotation, and retrieval \cite{ref9,ref10,ref11,ref12,refMS4,refMS5,refMS6}.

\subsection{Segmentation}

The process of segmentation divides an image in to multiple non-overlapping regions using a set of rules or criterion such as a set of similar pixels or intrinsic features such as color, contrast and texture \cite{ref14}. Segmentation reduces the search area in an image by dividing the original image into two classes such as object or background. The key aspect of image segmentation is to represent the image in a meaningful form such that it can be conveniently utilized and analyzed. The meaningful information extracted using the segmentation process in medical images involves shape, volume, relative position of organs, and abnormalities \cite{ref35,ref36}. In \cite{ref37}, an iterative 3D multi-scale Otsu thresholding algorithm is presented for the segementation of medical images. The effects of noise and weak edges are eliminated by representing images at multiple levels. In \cite{ref38}, a hybrid algorithm is proposed for an automatic segmentation of ultrasound images. The proposed method combine information from spatial constraint based kernel fuzzy clustering and distance regularized level set (DRLS) based edge features. Multiple experiments are conducted for evaluating the method on real as well as synthetically generated ultrasound images. A segmentation approach for 3D medical images is presented in \cite{ref39}, in which the system is capable of assessing and comparing the quality of segmentation. The approach is mainly based on the statistical shape based features coupled with extended hierarchal clustering algorithm and three different datasets of 3D medical images are used for experimentation. An expectation maximization approach is used for tumor segmentation on brain tumor image segmentation (BRATS) 2013 dataset. The method achieves considerable performance, but is only tested on a few images from the dataset and is not shown to generalize for all images in the dataset \cite{refSo}.

\subsection{Detection and Classification of Abnormality}
Abnormality detection in medical images is the process of identifying a certain type of disease such as tumor. Traditionally, clincial experts detect abnormalities, but it requires a lot of human effort and is time consuming. Therefore, development of automated systems for detection of abnormalities is gaining importance. Different methods are presented in literature for abnormality detection in medical images. In \cite{ref40}, an approach is presented for detection of the brain tumor using MRI segmentation fusion, namely potential field segmentation. The performance of this system is tested on a publicly available MRI benchmark, known as brain tumor image segmentation. A particle swarm optimization based algorithm for detection and classification of abnormalities in mammography images is presented in \cite{ref41},  which uses texture features and a support vector machine (SVM) based classifier. In \cite{ref42}, a method is presented for detection of myocardial abnormalities using cardiac magnetic resonance imaging.  

\subsection{Computer Aided Detection or Diagnosis }
A computer aided diagnosis (CAD) system is used in radiology, which assists the radiologist and clinical practitioners in interpreting the medical images. The system is based on algorithms which use machine learning, computer vision and medical image processing. In clinical practice, a typical CADx system serves as a second reader in making decisions that provides more detailed information about the abnormal region. A typical CADx system consists of the following stages, pre-processing, feature extraction, feature selection and classification \cite{ref43}. In literature, there are methods proposed for the diagnosis of diseases such as fatty liver \cite{ref45}, prostate cancer \cite{ref43}, dry eye \cite{ref47}, Alzheimer \cite{ref48}, and breast cancer \cite{ref50}. In \cite{refGAS}, hybrid  features are used for the detection glaucoma in fundus images. The optic disc is localized by employing support vector machine trained using local features extracted from the vessels \cite{refGAS1}. A hybrid of clinical and image based features are used for multi-class classification of alzheimer disease using the alzheimer disease neuro-image initiative (ADNI) dataset with reasonable accuracy \cite{refToo}.

\subsection{Medical Image Retrieval} 
Recent years have witnessed a broad use of computers and digital information systems in hospitals. The picture archiving and communication systems (PACSs) are producing large collections of medical images \cite{ref52,ref53,ref54}. The hospitals and radiology departments are producing a large number of medical images, ultimately resulting in huge medical image repositories. An automatic medical image classification and retreival system is required to efficiently deal with this big data. A speciliazed medical image retrieval system could assist the clinical experts in making a critical decision in disease prognosis and diagnosis. A timely and accurate deceison regarding the diagnosis of a patient's disease and its stage can be mabe by using similar cases retrieved by the reterival system \cite{ref55}. Text based and content based image retrieval (CBIR) methods have been commonly used for medical image retrieval. Text based retrieval methods were initially proposed in 1970s \cite{ref52}, where images were manually annotated with a text based description. In case, the textual annotation is done efficiently, the performance of such systems is fast and reliable. The drawback of such systems is that they cannot perform well in un-annotated image databases. Image annotation is not only a subjective matter but also a time taking process \cite{ref56}. In CBIR methods, texture, color and shape based features are used for searching and retrieving images from large collections of data \cite{ref57}. 

A CBIR system based on line edge singular value pattern (LESVP) is proposed in \cite{ref58}. In \cite{ref59}, a CBIR system for skin lesion images using reduced feature vector, classification and regression tree is presented. In \cite{ref55}, an Bag of Visual Words (BoVWs) approach is used along with scale invariant feature transform (SIFT) for the diagnosis of Alzheimer disease (AD). In \cite{ref60}, a supervised learning framework is presented for biomedical image retrieval, which uses the predicted class label from classifier for retrieval. It also uses image filtering and similarity fusion and multi-class support vector machine classifier. The use of class prediction eliminates irrelevant images and results in reducing the search area for similarity measurement in large databases \cite{ref61}.

\subsection{Evaluation Metrics for Medical Image Analysis System}

A typical medical image analysis system is evaluated by using different key performance measures such as accuracy, F1-score, precision, recall, sensitivity, specificity and dice coefficient. Mathematically, these measures are calculated as,
\begin{equation}
F1_{score} = 2 \times\frac{(Precision \times Recall)}{(Precision + Recall)},                                 
\end{equation}
where,
\begin{equation}
Precision=  \frac{TP}{(TP+FP)},
\end{equation}
and
\begin{equation}
Recall=  \frac{(TP)}{(TP+TN)},                                                                
\end{equation}
\begin{equation}
Accuracy =  \frac{(TP+TN)}{(TP+TN+FP+FN)},			       
\end{equation}
\begin{equation}
Sensitivity = \frac{TP}{(TP+FN)},
\end{equation}
\begin{equation}
Specificity =  \frac{TN}{(TN+FP)},
\end{equation}
\begin{equation}
Dice Score =\frac{2 \times |P \cap GT|}{|P|+|GT|},
\end{equation}
where true positive (TP) represents number of cases correctly recognized as defected, false positive (FP) represents number of cases incorrectly recognized as defected, true negative (TN) represents number of cases correctly recognized as non-defected and false negative (FN) represents number of cases incorrectly recognized as non-defected. In Eq. 7, P denotes the prediction as given by the system being evaluated for a given testing sample and GT represents the ground truth of the corresponding testing sample.

\section{Convolutional Neural Networks (CNNs)}\label{sec:dl}
 
Deep learning is a tool used for machine learning, where multiple linear as well as non-linear processing units are arranged in a deep architecutre to model high level abstraction present in the data \cite{ref62}. There are numerous deep learning techniques currently used in a variety of applications. These include auto-encoders, stacked auto-encoders, restricted Boltzmann machines (RBMs), deep belief networks (DBNs) and deep convolutional neural networks (CNNs). In recent years, CNN based methods have gained more popularity in vision systems as well as medical image analysis domain \cite{refMS1,refMS2,refMS3}. 


\begin{figure}[t]
\begin{center}
\begin{tabular}{c}
\includegraphics[width=120mm]{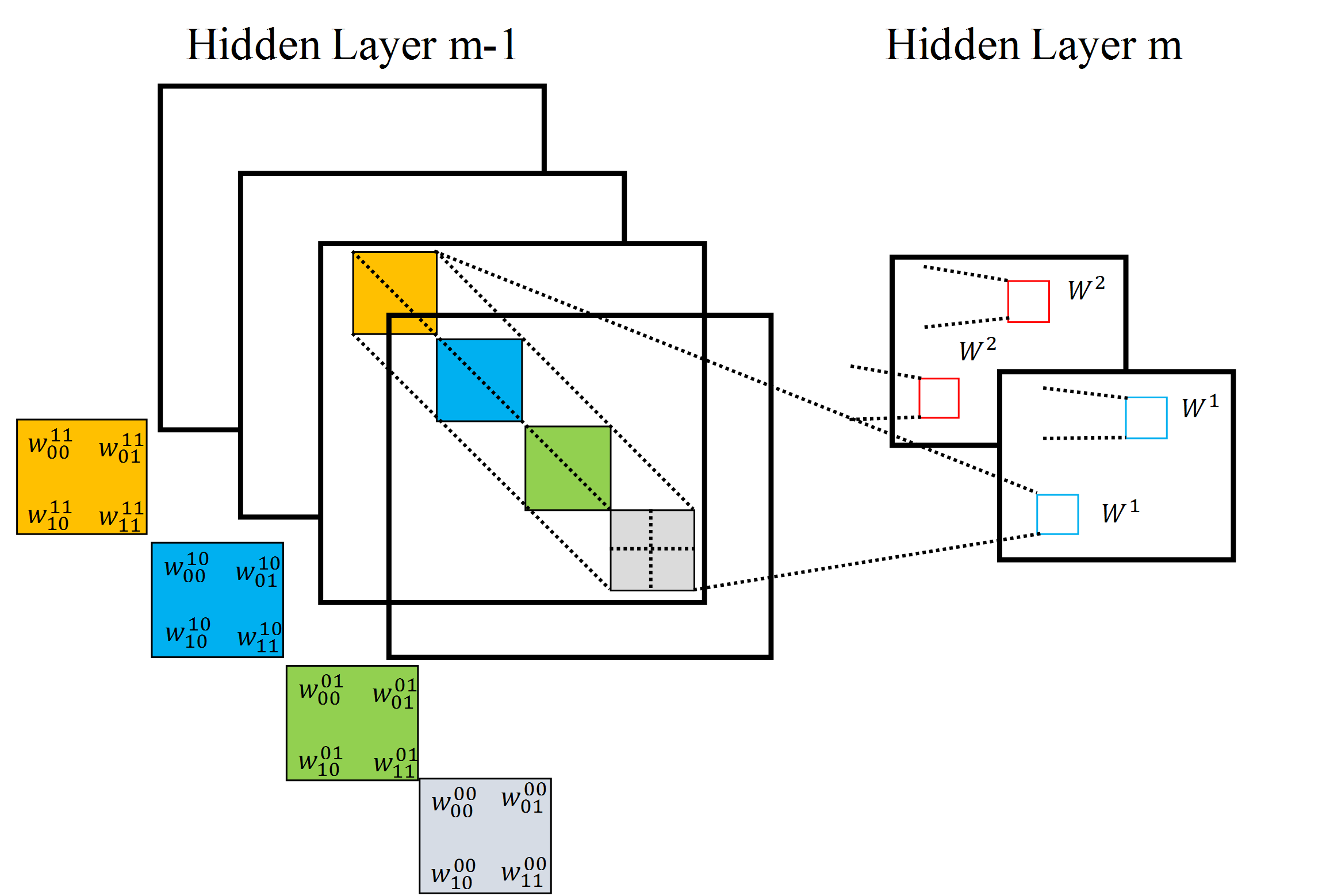}
\end{tabular}
\end{center}
\caption
{ \label{fig:fig2}
{Hidden layers in a convolutional neural network.}}
\end{figure}

CNNs are biologically inspired variants of multi-layer perceptrons. They tend to recognize visual patterns, directly from raw image pixels. In some cases, a minimal pre-processing is performed before feeding images to CNNs. These deep networks look at small patches of the input image, called receptive fields, by using multiple layer neurons and use shared weights in each convolutional layer. CNNs combine three architectural ideas for ensuring invariance for scale, shift and distortion to some extent. The first CNN model (LeNet-5) that was proposed for recognizing hand written characters is presented in \cite{ref73}. The local connections of patterns between the neurons of adjacent layers of CNN i.e., inputs from hidden units of a layer $m$ are taken as a subset of units in the layer $m-1$, units having spatially adjacent receptive fields for exploiting the spatial local correlation. Additionally, in CNN each filter $h_i$ is replicated around the whole visual field. These filters share bias and weight vectors to create a feature map. The gradient of shared weights is equal to the sum of gradients of the shared parameters. When convolution operation is performed on sub-regions of the whole image, a feature map is obtained. The process involves convolution of the input image or feature map with a linear filter with the addition of a bias followed by an application of a non-linear filter. A bias value is added such that it is independent of the output of previous layer. The bias values allow us to shift the activation function of a node in either left or right direction. For example, for a sigmoid function, the weights control the steepness of the output, whereas bias is used to offset the curve and allow better fitting of the model. The bias values are learned during the training model and allows an independent variable to control the activation. At a given layer, the $k^{th}$  filter is denoted symbolically as $h^k$, and the weights $W^k$ and bias $b_k$ determines their filters. The mathematical expression for obtaining feature maps is given as, 
\begin{equation}
h_{ij}^{k} = tanh((W^k * x)_{ij}+ b_k) ,                                  
\end{equation}
where, $tanh$ represents the tan hyperbolic function, and $*$ is used for the convolution operation. \Fig{fig2} illustrates two hidden layers in a CNN, where layer $m-1$ and $m$ has four and two features maps respectively i.e., $h^0$ and $h^1$ named as $w^1$ and $w^2$. These are calculated from pixels (neurons) of layer $m-1$ by using a $2\times 2$ window in the layer below as shown in \Fig{fig2} by the colored squares. The weights of these filter maps are 3D tensors, where one dimension gives indices for input feature maps, while the other two dimensions provides pixel coordinates. Combining it all together, $W_{ij}^{kl}$ represents the weight connected to each pixel of $k^{th}$ feature map at a hidden layer $m$ with $i^{th}$  feature map of a hidden layer $m-1$ and having coordinates $i,j$. 

Each neuron or node in a deep network is governed by an activation function, which controls the output. There are various activation functions used in deep learning literature such as linear, sigmoid, tanh, rectified linear unit (ReLU). A broader classification is made in the form of linear and non-linear activation function. A linear function passes the input at a neuron to the output without any change. Since, deep network architectures are designed to perform complex mathematical tasks, non-linear activation functions have found wide spread success. ReLU and its variations such as leaky-ReLU and parametric ReLU are non-linear activations used in many deep learning models due to their fast convergence characteristic. Pooling is another important concept in convolutional neural networks, which basically performs non-linear down sampling. There are different types of pooling used such as stochastic, max and mean pooling. Max pooling divides the input image into non-overlapping rectangular blocks and for every sub-block local maxima is considered in generating the output. Max pooling provides benefits in two ways, i.e., eliminating minimum values reduces computations for upper layers and it provides translational invariance. Concisely, it provides robustness while reducing the dimension of intermediate feature maps smartly. On the other hand, mean pooling replace the underlying block with its mean value. In stochastic pooling the activation function within the active pooling region is randomly selected. In addition to down-sampling the feature maps, pooling layers allows learning features for translational and rotational invariant classification \cite{lecun2015deep}. The pooling operation can also be performed on overlapping regions. In circumstances where weak spatial information surrounding the dominant regions of an image is also useful, fractional or overlapping regions for pooling could be beneficial \cite{ding2015deep}. 

There are various techniques used in deep learning to make the models learn and generalize better. This could include L1, L2 regularizer, dropout and batch normalization to name a few. A major issue in using deep convolutional network (DCNN) is over-fitting of the model during training. It has been shown that dropout is used successfully to avoid over-fitting \cite{srivastava2014dropout}. A dropout layer drops certain unit connections which are selected randomly. Dropout layer is widely used for regularization. In addition to dropout, batch normalization has also been successfully used for the purpose of regularization. The input data is divided into mini batches. It is shown that using batch normalization not only speeds up the training but, in some cases, preform regularization eliminating the need for using dropout layers \cite{ioffe2015batch}. The performance of a deep learning method is highly dependent on the data. In cases, where the availability of data is limited, various augmentation techniques are utilized \cite{KOOI2017303}. This may include random cropping, colour jittering, image flipping and random rotation \cite{perez2017effectiveness}.   

\section{Medical Image Analysis using CNN}

There is a wide variety of medical imaging modalities used for the purpose of clinical prognosis and diagnosis and in most cases the images look similar. This problem is solved by deep learning, where the network architecture allows learning difficult information. Hand crafted features work when expert knowledge about the field is available and generally make some strict assumptions. These assumptions may not be useful for certain tasks such as medical images. Therefore, with the hand-crafted features in some applications, it is difficult to differentiate between a healthy and non-healthy image. A classifier such as SVM does not provide an end to end solution. Features extracted form techniques such as scale invariant feature transform (SIFT) etc. are independent of the task or objective function in hand. Afterwards, sample representation is taken in term of bag of words (BOW), Fisher vector or some other mechanism. The classifier like SVM is applied on this representation and there is no mechanism for the of loss to improve local features as the process of feature extraction and classification is decoupled from each other. 

On the other hand, a DCNN learn features from the underlying data. These features are data driven and learnt in an end to end learning mechanism. The strength of DCNN is that the error signal obtained by the loss function is used/propagated back to improve the feature (the CNN filters learnt in the initial layers) extraction part and hence, DCNN results in better representation. The other advantage is that in the initial layers a DCNN captures edges, blobs and local structure, whereas the neurons in the higher layers focus more on different parts of human organs and some of the neurons in the final layers can consider whole organs.

\begin{figure}[t]
\begin{center}
\begin{tabular}{c}
\includegraphics[width=120mm]{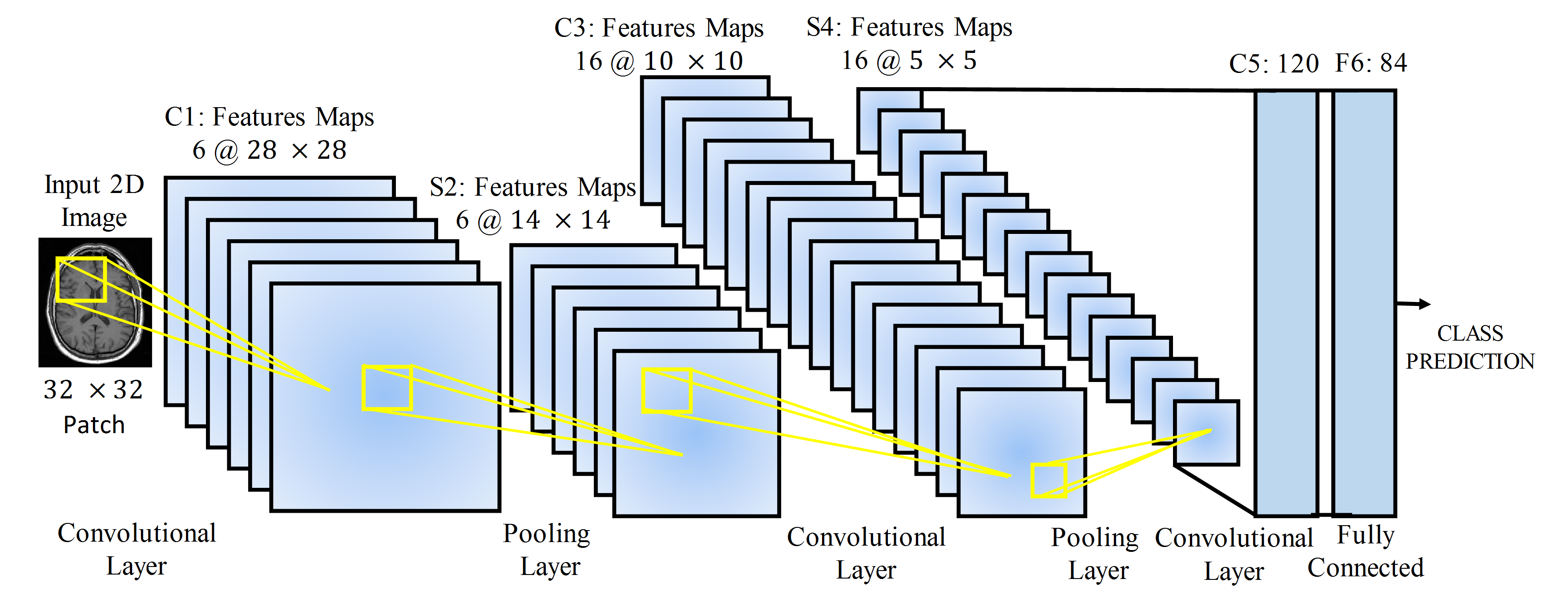}
\end{tabular}
\end{center}
\caption
{ \label{fig:fig3}
{A typical convolutional neural network architecture for medical image classification.}}
\end{figure}

\Fig{fig3} shows a CNN architecture like LeNet-5 for classification of medical images having $N$ classes accepting a patch of $32 \times 32$ from an original 2D medical image. The network has convolutional, max pooling and fully connected layers. Each convolutional layer generates a feature map of different size and the pooling layers reduce the size of feature maps to be transferred to the following layers. The fully connected layers at the output produce the required class prediction. The number of parameters required to define a network depends upon the number of layers, neurons in each layer, the connection between neurons. The training phase of the network makes sure that the best possible weights are learned, that would give high performance for the problem at hand. The advancement in deep learning methods and computational resources has inspired medical imaging researchers to incorporate deep learning in medical image analysis. Some recent studies have shown that deep learning algorithms are successfully used for medical image segmentation \cite{refS}, computer aided diagnosis \cite{ref95,ref96,ref97}, disease detection and classification \cite{ref74,ref90,ref91,ref92} and medical image retrieval \cite{ref98,ref99}. 

A deep learning based approach has been presented in \cite{ref81}, in which the network uses a convolutional layer in place of a fully connected layer to speed up the segmentation process. A cascaded architecture has been utilized, which concatenates the output of the first network with the input of succeeding network. The network presented in \cite{ref82} uses small kernels to classify pixels in MR image. The use of small kernels decreases network parameters, allowing to build deeper networks, without worrying about the dangers of over-fitting. Data augmentation and intensity normalization have been performed in pre-processing step to facilitate training process. Another CNN for brain tumor segmentation has been presented in \cite{ref83}. The architecture uses dropout regularizer to deal with over-fitting, while max-out layer is used as activation function. A two path eleven layers deep convolutional neural network has been presented in \cite{ref84} for brain lesion segmentation. The network is trained using a dense training method using 3D patches. A  3D fully connected conditional random field has been used to remove false positives as well as to perform multiple predictions. The CNN based method presented in \cite{ref85} deals with the problem of contextual information by using a global-based method, where an entire MRI slice is taken into account in contrast to patch based approach. A re-weighting training procedure has been used to deal with the data imbalance problem. A 3D convolutional network for brain tumor segmentation for the BRATS challenge has been presented in \cite{ref86}. The network uses a two-path approach to classify each pixel in an MR image. In \cite{refS}, a deep convolutional neural network is presented for brain tumor segmentation, where a patch based approach with inception method is used for training purpose. Drop-out, batch normalization and inception modules are utilized to build the proposed ILinear nexus architecture. The problem of over-fitting, which arises due to scarcity of data, is removed by using drop-out regularizer. \Tab{tab1} highlights the usage of CNN based architectures for segmentation of medical images. 


\begin{table}
\caption{The application of CNN based methods for medical image segmentation.}\label{tab:tab1}
\centering
\scalebox{1}{
\begin{tabular}{|c|c|c|c|c|}

\hline
 Method    &  Dataset & \multicolumn{3}{c}{Dice} \vline \\ \cline{3-5}
   &   & Complete & Core & Enhancing\\ \hline
\thead{ InputCascade\\ CNN \cite{ref81}} & 		\thead{ BRATS\\ 2013}		& 	0.88		& 	0.79		& 	0.73 \\
\hline
\thead{ Pereira \cite{ref82}} 		      &		\thead{ BRATS\\ 2013}	&	0.84		&	0.72		&	0.62 \\
\hline
\thead{Lisa \cite{ref83}} 		      &		\thead{ BRATS\\ 2013}	&	0.79		&	0.68		&	0.57 \\
\hline
\thead{Deep \\medic \cite{ref84}} 	      &           \thead{ BRATS\\ 2015}	           &        0.89		& 	0.75		&	0.72 \\
\hline
\thead{SegNet \cite{ref85}} 		      & 		\thead{ BRATS\\ 2015}	& 	0.85		& 	0.68		&	0.68 \\
\hline
\thead{3DNet 3 \cite{ref86}} 	      & 		\thead{ BRATS\\ 2015}		& 	0.91		& 	0.83		&	0.76 \\
\hline
\thead{Cascaded  \\ neural  networks \cite{refS}} & 	\thead{ BRATS\\ 2015}	&  	0.86		& 	0.87		& 	0.90 \\
\hline

\end{tabular}
}
\end{table}

\begin{table}
\caption{Some recent clinical applications of CNN based methods.}\label{tab:tab2}
\centering
\scalebox{0.77}{
\begin{tabular}{|c|c|c|c|c|}

\hline
 Application	 &	 Method	& 	Dataset	&	 \thead{Number of images/ \\ classes}  & Accuracy   \\   \hline

\thead{ Body Part \\ Recognition} & 	\thead{Two Stage \\ Convolutional \\ Neural Network \cite{ref91}}	& \thead{ CT Slices of \\ 12 body organs}   & \thead{6000 Synthetic \\ 7489 Transversal slices \\12 classes} & 92.23\% \\ \hline

\thead{ Lung Texture\\ Classification and \\Airway Detection} 	& \thead{ Convolutional Restricted \\ Blotzman Machine \cite{ref90} } 	& \thead{ILD (interstitial lung\\ diseases ) CT scans} & \thead{ 73 CT scans \\ 5 classes} & ~89\% \\ \hline

 \thead{ Lung Pattern \\ Classification} 	& \thead{ Convolutional\\ Neural Network \cite{ref74} }	& \thead{ILD (interstitial lung \\ diseases )\\ CT scans}&	\thead{109 high resolution \\ CT scans \\7 classes} & 85.5\% \\ \hline
 
\thead{ Detection and \\ Classification \\ of Nuceli}	& \thead{Two architectures of\\ Convolutional\\ Neural Network\cite{ref92}}	& \thead{histology images of \\ colorectal \\ adenocarcinomas} & \thead{100 \\ histology images \\ 4 classes} & ~80.2\% \\ \hline

\thead{Thyroid \\ Nodule \\ Diagnosis} 	& \thead{Pre-Trained \\Convolutional \\Neural Network \cite{ref95}}	&  \thead{Ultrasound \\ Images}  & \thead{15,000 \\ ultrasound images \\2 classes} & 	~83\% \\ \hline

\thead{ Breast \\Cancer \\ Diagnosis}  & \thead{ Convolutional Neural \\ Network using \\ semi supervised \\ learning \cite{ref96}} &  \thead{Mammographic Images \\ with ROIs } & \thead{3158 ROIs \\2 classes} & 82.43\% \\  \hline

\thead{Diabetic \\ Retinopathy }	& \thead{ Convolutional \\Neural Network \cite{ref97}}	& \thead{Kaggle \\ Dataset} &	\thead{80,000 images \\5 classes} & 75\% \\ \hline

\thead{Medical Image \\ Classification\\ and Retrieval} 	&  \thead{Convolutional\\ Neural Network \cite{ref98}}	&  \thead{ Multimodal Dataset \\ having 24 classes}   & \thead{7200 multi-modal \\ images \\24 classes} & \thead{99.77\% \\  } \\ \hline 

\thead{Radiographic \\ Image Retrieval} & 	\thead{Convolutional Neural \\ Network \cite{ref99}}	& \thead{ IRMA \\Database }	& \thead{14,410 images \\31 classes }& 97.79\% \\ \hline

\thead{Multi-class \\ Classification\\ of Alzheimer Disease} 	&  \thead{Convolutional\\ Neural Network \cite{refA1}}	&  \thead{ ADNI \\ database}  & \thead{38,024 images \\ 4 classes} & \thead{98.88\% \\  } \\ \hline

\end{tabular}
}
\end{table}

A method for classification of lung disease using a convolutional neural network is presented in \cite{ref74}, which uses two databases of interstitial lung diseases (ILDs) and CT scans each having a dimension of $512\times 512$. A total of $14696$ image patches are derived from the original CT scans and used to train the network. A method based on convolutional classification restricted Boltzmann machine for lung CT image analysis is presented in \cite{ref90}. Two different datsets containing lung CT scans are used for classification of lung tissue and detection of airway center line. The network is trained on $32\times 32$ image patches selected along a gird with a 16-voxel overlap. A patch is retained if it has $75\%$ of voxel belonging to the same class. In \cite{ref91}, a framework for body organ recognition is presented based on two-stage multiple instance deep learning. In the first stage, discriminative and non-informative patches are extracted using CNN. In the second stage, fine tuning of the network parameters is performed on extracted discriminative patches. The experiments are conducted for the classification of synthetic dataset as well as the body part classification of 2D CT slices. In \cite{ref92}, a locality sensitive deep learning algorithm called spatially constrained convolutional neural networks is presented for the detection and classification of the nucleus in histological images of colon cancer. A novel neighboring ensemble predictor is proposed for accurate classification of nuclei and is coupled with CNN. A large dataset having $20,000$ annotated nuclei of four classes of colorectal adenocarcinoma images is used for evaluation purposes. In \cite{ref98}, a deep convolutional neural network has been proposed to retrieve multimodal images. An intermodal dataset having five modalities and twenty-four classes are used to train the network for the purpose of classification. Three fully connected layers are used at the last part of the network for extracting features, which are use for the retrieval. A content based medical image retrieval (CBMIR) system based on CNN for radiographic images is proposed in \cite{ref99}. Image retrieval in medical application (IRMA) database is used for the evaluation of the proposed CBMIR system. In \cite{ref96}, a hybrid thyroid module diagnosis system has been proposed by using two pre-trained CNNs. The models differs in terms of the number of convolutional and fully connected layers. A soft-max classifier is used for diagnosis and results are validated on $15000$ ultrasound images. A semi-supervised deep CNN based learning scheme is proposed for the diagnosis of breast cancer\cite{ref97}, and is trained on a small set of labeled data. In \cite{ref98}, a CNN based approach is proposed for diabetic retinopathy using colored fundus images. The network classify the images into three classes i.e., aneurysms, exudate and haemorrhages and also provide the diagnosis. The proposed architecture is tested on dataset comprising of 80000 images. In \cite{refA1,refA2}, deep neural network including GoogLeNet and ResNet are successfully used for multi-class classification of Alzheimer's  disease patients using the ADNI dataset. An accuracy of $98.88\%$ is achieved, which is higher than the traditional machine learning approaches used for Alzheimer's disease detection.

\begin{table}
\caption{A comparison of methods used for ILD classification.}\label{tab:tab3}
\centering
\scalebox{0.75}{
\begin{tabular}{|c|c|c|c|c|c|}

\hline
Method	& Features	& Classifier	& Precision	& Recall	& F1 Score    \\   \hline
\multirow{5}{*}{Yan et al. \cite{ref91}} & \thead{Bag of Words (BoW) \\ + \\SIFT}	& Linear regression (LR)	&62.21	&63.37	&62.78 \\ \cline{2-6}
& \thead{BoW + SIFT \\}	& SVM	&63.72	&64.63	&64.17 \\ \cline{2-6}
&\thead{Histogram of oriented \\ gradients (HOG)}	& LR	&67.74	&68.71	&68.22 \\ \cline{2-6}
&HOG	&\thead{SVM	\\ }&76.39	&76.75	&76.57 \\ \cline{2-6}
& CNN	& \thead{ CNN \\ } & 92.25	& 92.21	& 92.23 \\ \hline

\end{tabular}
}
\end{table}

\begin{table}
\caption{A comparison of CNN based method with other state-of-the-art methods for body organ recognition.}\label{tab:tab4}
\centering
\scalebox{0.8}{
\begin{tabular}{|c|c|c|c|c|}

\hline
Method	& Features	& Classifier	& $F_{avg}$	& Accuracy    \\  \hline
Gangeh \cite{ref100} & \thead{Intensity \\ Texon}	 & SVM-radial basis function	 &0.7127	 &0.7152 \\ \hline
Sorensen \cite{ref101} & \thead{ Local binary pattern + \\ Histogram}	 & K-nearest neighbour	 &0.7322	 &0.7333 \\ \hline
Anthimopoulous \cite{ref102} & \thead{ Local Discrete Cosine transform + \\Histogram} &	Random forest	&0.7786 &	0.7809 \\ \hline
Anthimopoulous \cite{ref74}  &	 & \thead{CNN \\ }	 &0.8547	 &0.8561 \\ \hline

\end{tabular}
}
\end{table}

\Tab{tab2} highlights CNN applications for the detection and classification task, computer aided diagnosis and medical image retrieval. It is seen that CNN based networks are successful in application areas dealing with multiple modalities for various tasks in medical image analysis and provide promising results in almost every case. The results can vary with the number of images used, number of classes, and the choice of the DCNN model. Looking at these successes of CNN in medical domain, it seems that convolutional networks will play a crucial role in the development of future medical image analysis systems. Deep convolutional neural networks have proven to give high performance in medical image analysis domain when compared with other techniques applied in similar application areas. \Tab{tab3}, summarises results of different techniques used for lung pattern classification in ILD disease. The CNN based method outperforms other methods in major performance indicators. \Tab{tab4} shows a comparison of the performance of a CNN based method and other state-of-the-art computer vision based methods for body organ recognition. It is evident that the CNN based method achieves significant improvement in key performance indicators. 

\section{Discussion}

In this section, various considerations for adopting deep learning methods in medical image analysis are discussed. A roadmap for the future of artificial intelligence in medical image analysis is also drawn in the light of recent success of deep learning for these tasks. 

\subsection{Various Deep Learning Architectures for Medical Image Analysis}

The success of convolutional neural networks in medical image analysis is evident from a wide spectrum of literature that is recently available \cite{chen2017deep}. There are multiple CNN architectures reported in literature to deal with different imaging modalities and tasks involved in medical image analysis \cite{refS} - \cite{refA1}. These architectures include conventional CNN, multiple layer networks, cascaded networks, semi- and fully supervised training models and transfer learning. In most cases, the data available is limited and expert annotations are scarce. In general, shallow networks have been preferred in medical image analysis, when compared with very deep CNNs employed in computer vision applications \cite{hoo2016deep} \cite{simonyan2014very}. In \cite{10.1007/978-3-319-46723-8_49}, a U shaped network is used for the purpose of semi-automated segmentation of sparsely annotated volumetric data. This architecture introduces skip connections and use convolution, deconvolution in a structured manner.  A modification to U-Net is proposed in \cite{zhou2018unet++}, which is applied on a variety of medical datasets for segmentation tasks. In \cite{chen2018w}, a W-shaped network is proposed for 2D medical image segmentation task. In \cite{7785132}, a volumetric solution is proposed for end to end segmentation of prostate cancer. A convolutional-deconvolutional network based on a capsule architecture is proposed in \cite{lalonde2018capsules} for lung image segmentation and is shown to substantially reduce the number of parameters required when compared to U-Net architecture. This analysis shows that different DCNN network architectures are adopted or proposed for medical image analysis. These architectures focus on reducing the parameter space, improve computation time, and handle 3D data. It is generally found that DCNN based architectures have found wider success in dealing with medical image data, when compared to other deep learning frameworks.    

\subsection{3D Imaging Modalities}
A large amount of data produced in the medical domain has 3-dimensional information. This is particularly true for volumetric imaging modalities such as CT and MRI. Medical image analysis can benefit from this enriched information. Deep learning methods generally adopt different methods to handle this 3D information. This can involve converting 3D volume data into 2D slices and combination of features from 2D and multi-view planes to benefit from the contextual information \cite{chen2016voxresnet} \cite{setio2016pulmonary}. Recent techniques are proposed using 3D CNN to fully benefit from the available information \cite{brosch2016deep} \cite{cciccek20163d}. In \cite{ceschin2018computational}, a fully 3D DCNN is used for the classification of dysmaturation in neonatal MRI image data. In \cite{ghafoorian2017deep}, a two stage network is used for the detection of vascular origin lacunes, where a fully 3D CNN used in the second stage. The performance of the system is close to trained raters. In \cite{meijs2018artery}, a 3D CNN is used for the segmentation of cerebral vasculature using 4D CT data. In \cite{kamnitsas2017efficient}, brain lesion segmentation is performed using 3D CNN. A 3D fully connected conditional random field (CRF) is used for post processing. A geometric CNN is proposed in \cite{seong2018geometric} to deal with geometric shapes in medical imaging, particularly targeting brain data. The utilization of 3D CNN has been limited in literature due to the size of network and number of parameters involved. This also leads to slow inference due to 3D convolutions. A hybrid of 2D/3D networks and the availability of more compute power is encouraging the use of fully automated 3D network architectures.         

\subsection{Limitation of Deep Learning and Future Prospects}
Despite the ability of deep learning methods to give better or higher performance, there are some limitations of deep learning techniques, which could limit their application in clinical domain. Deep learning architecture requires a large amount of training data and computational power. A lack in computational power will lead to a need for more time to train the network, which would depend on the size of training data used. Most deep learning techniques such as convolutional neural network requires labelled data for supervised learning and manual labelling of medical images is a difficult task. These limitations are being overcome with every passing day due to the availability of more computation power, improved data storage facilities, increasing number of digitally stored medical images and improving architecture of the deep networks. The application of deep learning in medical image analysis also suffers from the black box problem in AI, where the inputs and outputs are known but the internal representations are not very well understood. These methods are also affected by noise and illumination problems inherent in medical images. The noise can be removed using pre-processing steps to improve the performance \cite{refS}. 

A possible solution to deal with these limitations is to use transfer learning, where a pre-trained network on a large dataset (such as ImageNet) is used as a starting point for training on medical data. This typically includes reducing the learning rate by one or two orders of magnitude (i.e., if a typical learning rate is $1e-2$, reduce it to $1e-3$ or $1e-4$) and increase the local learning rate of the newly introduce layers by a factor of 10. Also, as an alternative the DCNN model can be pretrained by converting ImageNet data into gray scale images. However, it may require more computation resources (such as GPUs) to train on the whole ImageNet data. The best option would be to train DCNN model on large scale annotated medical image data. This underlying task for pre-training can be as simple as organ classification \cite{ref98} or binary classification task of benign or malignant images. Different modalities e.g., X-ray, MRI, and CT can be combined for this task. This pre-trained model can be used in transfer learning for fine tuning a network for a particular problem at hand. 

In general, shallow networks are used in situations where data is scarce. One of the most important factors in deep learning is the training data. However, this is partially addressed by using transfer learning. However, even in the presence of transfer learning more data on the target domain will give better performance. The use of generative adversarial network (GAN) \cite{tzeng2017adversarial} can be explored in the medical imaging field in cases where the data is scarce. One of the main advantages of transfer learning is to enable the use of deeper models to relatively small dataset. In general, a deeper DCNN architecture is the better for the performance.

\section{Conclusion}\label{sec:conc}
A comprehensive review of deep learning techniques and their application in the field of medical image analysis is presented. It is concluded that convolutional neural network based deep learning methods are finding greater acceptability in all sub-fields of medical image analysis including classification, detection, and segmentation. The problems associated with deep learning techniques due to scarce data and limited labels is addressed by using techniques such as data augmentation and transfer learning. For larger datasets, availability of more compute power and better DL architectures is paving the way for a higher performance. This success would ultimately translate into improved computer aided diagnosis and detection systems. Further research is required to adopt these methods for those imaging modalities, where these techniques are not currently applied. The recent success indicates that deep learning techniques would greatly benefit the advancement of medical image analysis. 


\bibliographystyle{elsarticle-num}
\bibliography{Reference}   

\end{document}